\documentclass[10pt,twocolumn,letterpaper]{article}

\usepackage{iccv}              
\usepackage{times}
\usepackage{epsfig}
\usepackage{graphicx}
\usepackage{amsmath}
\usepackage{amssymb}
\usepackage{pifont}
\usepackage{paralist}
\usepackage{enumitem}
\usepackage{makecell}
\usepackage{tabularx}
\usepackage{subcaption}

\usepackage{booktabs} 
\usepackage{multirow}
\usepackage{color, soul}
\usepackage[dvipsnames]{xcolor}

\usepackage{float}
\usepackage{stfloats}
\usepackage{soul}
\usepackage{multirow}
\usepackage{colortbl}
\usepackage{underscore}

\definecolor{iccvblue}{rgb}{0.21,0.49,0.74}
\usepackage[pagebackref,breaklinks,colorlinks,allcolors=iccvblue]{hyperref}

\def\paperID{2750} 
\def\confName{ICCV}
\def\confYear{2025}

\title{Recognizing Actions from Robotic View for Natural Human-Robot Interaction}

\author{
    Ziyi Wang\textsuperscript{1}\quad 
    Peiming Li\textsuperscript{1}\quad 
    Hong Liu\textsuperscript{1}\quad  
    Zhichao Deng\textsuperscript{2}\quad 
    Can Wang\textsuperscript{3}\\
    Jun Liu\textsuperscript{4}\quad  
    Junsong Yuan\textsuperscript{5}\quad 
    Mengyuan Liu\textsuperscript{1$\dagger$}\\
    \small
    $^1$State Key Laboratory of General Artificial Intelligence, Peking University, Shenzhen Graduate School\\
    \small
    $^2$Sun Yat-sen University\quad
    $^3$Kiel University \quad
    $^4$Lancaster University \quad
    $^5$State University of New York at Buffalo \\
    {\tt \small
    \{ziyiwang,lipeiming1001\}@stu.pku.edu.cn\quad  
    hongliu@pku.edu.cn\quad
    dengzhch3@mail2.sysu.edu.cn
    }\\
    {\tt \small
    canwang@pku.edu.cn\quad  
    j.liu81@lancaster.ac.uk\quad  
    jsyuan@buffalo.edu\quad  
    liumengyuan@pku.edu.cn\quad  
    }
}

\begin{document}
\maketitle

\def\thefootnote{}\footnotetext{$\dag$ Corresponding author: Mengyuan Liu (liumengyuan@pku.edu.cn)} 

\begin{abstract}
Natural Human-Robot Interaction (N-HRI) requires robots to recognize human actions at varying distances and states, regardless of whether the robot itself is in motion or stationary. This setup is more flexible and practical than conventional human action recognition tasks. However, existing benchmarks designed for traditional action recognition fail to address the unique complexities in N-HRI due to limited data, modalities, task categories, and diversity of subjects and environments. To address these challenges, we introduce ACTIVE (Action from Robotic View), a large-scale dataset tailored specifically for perception-centric robotic views prevalent in mobile service robots. ACTIVE comprises 30 composite action categories, 80 participants, and 46,868 annotated video instances, covering both RGB and point cloud modalities. Participants performed various human actions in diverse environments at distances ranging from 3m to 50m, while the camera platform was also mobile, simulating real-world scenarios of robot perception with varying camera heights due to uneven ground. This comprehensive and challenging benchmark aims to advance action and attribute recognition research in N-HRI. Furthermore, we propose ACTIVE-PC, a method that accurately perceives human actions at long distances using Multilevel Neighborhood Sampling, Layered Recognizers, Elastic Ellipse Query, and precise decoupling of kinematic interference from human actions. Experimental results demonstrate the effectiveness of ACTIVE-PC. Our code is available at: \url{https://github.com/wangzy01/ACTIVE-Action-from-Robotic-View}.
\end{abstract}
    
\section{Introduction}
\label{sec:intro}
A robust Natural Human-Robot Interaction (N-HRI) system can provide a more personalized and intelligent interaction experience by adapting to changes in user needs and evolving action patterns. It has a wide range of applications, such as in service robots \cite{robotde}, public safety surveillance \cite{106}, and medical diagnostics \cite{Yenduri2022FinegrainedAR,health}. N-HRI requires robots to have the ability to adapt in real-time, not only understanding human actions but also considering changes in human position (\emph{e.g.}, walking, standing) and their own motion. Therefore, compared to traditional human action recognition, human action recognition in N-HRI involves handling more complex human-robot interaction scenarios.

\begin{figure}[!t]
	\centering 
	\begin{tabular}{c}		
		\includegraphics[width=8cm]{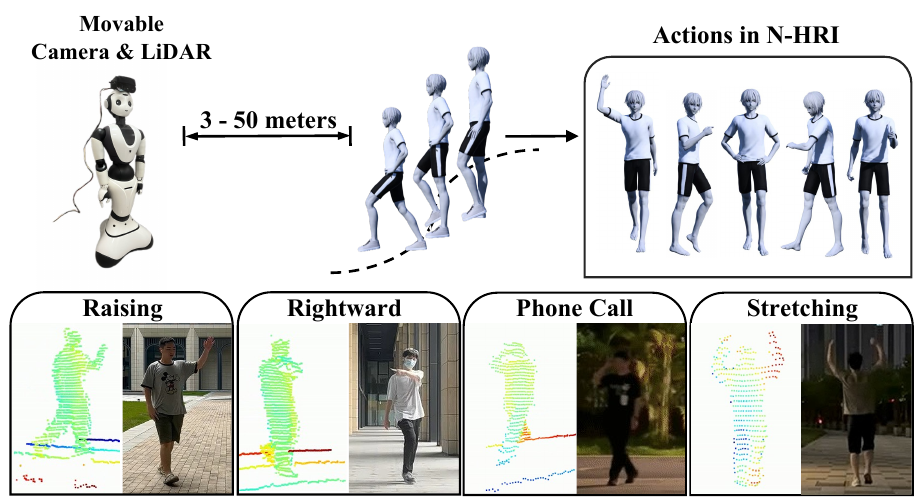}\\
	\end{tabular}%
    \vspace{-2mm}
	\caption{\textbf{Data Collection Process and Samples of the ACTIVE Dataset:} The movable camera and LiDAR capture human motion at distances ranging from 3 to 50 meters, with scene shaking introduced during the recording process. For visualization purposes, the samples have been scaled and cropped.}\label{fig:fig1}%
    \vspace{-6mm}
\end{figure}%

Compared to traditional human action recognition videos, videos captured from perception-centric robotic views in N-HRI scenarios exhibit diverse viewpoints, greater kinematic disturbances, and varied subject resolutions due to changing human-robot positions, robot movements, and uneven ground. While different from manipulator-mounted or end-effector perspectives common in industrial robots, this perception-centric viewpoint aligns closely with real-world applications, such as mobile assistants and patrol robots sensing humans at distances of 3–50 meters using forward-facing sensors. These challenges necessitate specialized methods tailored specifically to the unique characteristics of robotic views in N-HRI.

\begin{table*}
\caption{A comparison of the ACTIVE dataset, designed for N-HRI, with other widely used general action and micro-action datasets. ACTIVE is the first large-scale action recognition dataset specifically focused on N-HRI.
}
\vspace{-0.4cm}
\scriptsize
\setlength\tabcolsep{3.8pt}
\centering
\begin{tabular}{!{\vrule width 1.0pt}c|c!{\vrule width 1.0pt}|c|c|c|c|c!{\vrule width 1.0pt}|c|c|c!{\vrule width 1.0pt}|c|c!{\vrule width 1.0pt}}

\noalign{\hrule height 1.0pt}

\multicolumn{2}{!{\vrule width 1.0pt}c!{\vrule width 1.0pt}|}{} & 
\multicolumn{5}{c!{\vrule width 1.0pt}|}{\textbf{General Action}} & \multicolumn{3}{c!{\vrule width 1.0pt}|}{\textbf{Fine-Grained Action}} & \multicolumn{1}{c!{\vrule width 1.0pt}}{\textbf{Actions from Robotic View}}\\ 
\noalign{\hrule height 1.0pt}
\multicolumn{2}{!{\vrule width 1.0pt}c!{\vrule width 1.0pt}|}{Dataset Attribute} & 
\begin{tabular}[c]{@{}c@{}}MSR \\ Action3D\end{tabular} &
\begin{tabular}[c]{@{}c@{}}Multiview \\ 3D Event\end{tabular} &
\begin{tabular}[c]{@{}c@{}}NTU \\ RGB+D 120\end{tabular}
& \begin{tabular}[c]{@{}c@{}}UAV \\ Human\end{tabular} &
\begin{tabular}[c]{@{}c@{}} PA \\ HMDB51 \end{tabular} &
\begin{tabular}[c]{@{}c@{}} HOMAGE \end{tabular} &
\begin{tabular}[c]{@{}c@{}} iMiGUE\end{tabular} &  SMG  &

\multicolumn{1}{c!{\vrule width 1.0pt}}{\begin{tabular}[c!{\vrule width 1.0pt}]{@{}c@{}} \textbf{ACTIVE} \\ \textbf{(Ours)}\end{tabular}}  \\                    
\noalign{\hrule height 1.0pt}

\multirow{3}{*}{Action Recog.} & \# Videos & 567 & 3,815 & 114,480 & 67,428 & 515 & 1,752 & 18,499 & 3,712 & \textbf{46,868} \\ \cline{2-11} 

& \# Subjects & 20 & 10 & 106 & 119 & - & 27 & 72 & 40 & \textbf{80} \\ \cline{2-11} 

& \# Classes & 10 & 10 & 120 & 155 & 51 & 75 & 32 & 16 & \textbf{30} \\ \hline

Attribute Recog. & \# Samples & $\times$ & $\times$ & $\times$ & 22,476 & 515 & $\times$ & $\times$ & $\times$ & \textbf{46,868} \\ 

\noalign{\hrule height 1.0pt}

\multicolumn{2}{!{\vrule width 1.0pt}c!{\vrule width 1.0pt}|}{Data Modality} & D  & RGB+D & RGB+D+IR & RGB+D+IR &  RGB & RGB+IR & RGB & RGB+D & \textbf{RGB+Point Cloud} \\ \hline
 
\multicolumn{2}{!{\vrule width 1.0pt}c!{\vrule width 1.0pt}|}{Sensors}  
& \begin{tabular}[c]{@{}c@{}} Kinect\end{tabular}
& \begin{tabular}[c]{@{}c@{}} Kinect\end{tabular} 
& \begin{tabular}[c]{@{}c@{}} Kinect V2 \end{tabular} 
& \begin{tabular}[c]{@{}c@{}} Azure DK  
\end{tabular}
& NA & RGB Cam. & RGB Cam. & Kinect V2 
& \begin{tabular}[c]{@{}c@{}} \textbf{LiDAR, RGB Cam.} \end{tabular} \\

\noalign{\hrule height 1.0pt}

\multirow{2}{*}{\begin{tabular}[c]{@{}c@{}}Capturing \\ Scenarios\end{tabular}} & \# Sites & 1 & NA & 3 & 45 & NA & NA & NA & 1 & \textbf{6} \\ \cline{2-11}

& Outdoor & $\times$ & $\times$ & $\checkmark$ & $\checkmark$ & $\times$ & $\times$ & $\times$ & $\times$ & \textbf{$\checkmark$} \\ \hline

\multirow{1}{*}{\begin{tabular}[c]{@{}c@{}}Shooting Time\end{tabular}} & Night & $\times$ & $\times$ & $\times$ & $\checkmark$ & $\times$ & $\times$ & $\times$ & $\times$ & \textbf{$\checkmark$} \\  \hline

\multicolumn{2}{!{\vrule width 1.0pt}c!{\vrule width 1.0pt}|}{Composite Action} & $\times$ & $\times$ & $\times$ & $\times$ & $\times$ & $\checkmark$ & $\checkmark$ & $\checkmark$ & \textbf{$\checkmark$} \\ \hline

\multicolumn{2}{!{\vrule width 1.0pt}c!{\vrule width 1.0pt}|}{Dynamic Distance} & $\times$ & $\times$ & $\times$ & $\times$ & $\times$ & $\times$ & $\times$ & $\times$ & \textbf{$\checkmark$} \\ \hline

\multicolumn{2}{!{\vrule width 1.0pt}c!{\vrule width 1.0pt}|}{Platform Motion} & $\times$ & $\times$ & $\times$ & $\checkmark$ & $\times$ & $\times$ & $\times$ & $\times$ & \textbf{$\checkmark$} \\ 

\noalign{\hrule height 1.0pt}

\multicolumn{2}{!{\vrule width 1.0pt}c!{\vrule width 1.0pt}|}{Source} & CVPR & ICCV & TPAMI & CVPR & TPAMI & CVPR & CVPR &  IJCV & - \\ \hline

\multicolumn{2}{!{\vrule width 1.0pt}c!{\vrule width 1.0pt}|}{Year} & 2010 & 2013 & 2019 & 2021 & 2020 & 2021 & 2021 & 2023 & - \\ 

\noalign{\hrule height 1.0pt}

\end{tabular}
\vspace{-4mm}
\label{tab:dataset1}
\end{table*}

Existing work has already demonstrated the importance of using large, comprehensive, and challenging datasets to develop and evaluate state-of-the-art deep learning methods. However, in the field of human action understanding for N-HRI scenarios, current datasets \cite{cvpr10msraction, NTU} have limitations in several aspects, including: 1) Limited Sample Size: Large-scale datasets are often critical for mitigating overfitting and enhancing the generalization ability of models developed on them. 2) Fixed Capture Modes: Most existing datasets rely on fixed relative positions and backgrounds, with subjects typically standing still and performing predefined actions or postures. In contrast, humans in N-HRI are usually mobile. 3) Constrained Capture Scenes: In real-world applications, N-HRI needs to function in diverse environments (\emph{e.g.}, indoor, outdoor) and across various time periods (\emph{e.g.}, day, night). However, samples in current datasets are often collected under similar conditions, which oversimplifies the challenges in real-world N-HRI scenarios. 4) Limited Capture Distances: N-HRI often involves frequent changes in the human-robot position, leading not only to significant motion blur but also to considerable resolution changes. Yet, in most existing datasets, humans only exhibit slow and subtle movements, with fixed capture distances. 5) Limited Data Modalities: In real-world applications, different types of sensors are often deployed to collect data under varying conditions. For instance, LiDAR sensors can be used for long-distance N-HRI understanding, while RGB cameras are typically used for close-range scenes that require rich detail. This highlights the importance of collecting multimodal data to analyze human action under different conditions. However, most existing human action datasets only provide traditional RGB video samples or depth modalities. Devices like Kinect suffer from limited capture range and poor performance under daylight conditions. These limitations emphasize the need for a more robust and diverse dataset that better reflects the dynamic and real-world complexities of N-HRI scenarios.
\vspace{-2mm}

The aforementioned limitations of existing datasets clearly highlight the need for a larger, more challenging, and comprehensive dataset for human action analysis for N-HRI. To this end, we introduce ACTIVE, a large-scale multimodal dataset in this field. Specifically, ACTIVE includes 30 action categories with disturbing motion (\emph{e.g.}, Heading + Nodding), involving 80 participants and attribute information such as gender and clothing. It consists of a total of 46,868 video instances, captured using sensors including LiDAR and RGB cameras. During data capture, participants perform various human actions in different environments (indoor and outdoor, day and night) at varying capture distances (ranging from 3m to a long-range 50m). Additionally, we simulate the robot's motion by introducing movement to the camera platform. The use of different sensors allows our dataset to provide rich data modalities, including both RGB and point cloud data. Overall, ACTIVE serves as a more challenging dataset, encompassing action recognition video instances in N-HRI scenarios and introducing diverse kinematic disturbances, which can promote research on human action understanding in N-HRI settings. \cref{tab:dataset1} presents a statistical comparison between ACTIVE and existing datasets. More details about ACTIVE will be discussed in \cref{sec:3}.

\vspace{-2mm}
As shown in \cref{fig:fig1}, unlike general actions, ACTIVE incorporates variations in human-robot states (\emph{e.g.}, changes in relative distance and position through disturbing motion like walking and camera platform movement). Recognizing actions from robotic view aims to understand human actions in N-HRI scenarios, with the goal of filtering out the influence of kinematic disturbances from the robot's perspective and recognizing human actions to guide N-HRI. The main challenges of ACTIVE are as follows: 1) At long distances (\emph{e.g.}, 50 meters), human actions are subtle and rapid, making it even more difficult to capture the fine details of these actions. 2) In dynamic interaction scenarios, large-scale kinematic disturbances (such as the motion of the robot platform, environmental changes like lighting and obstacles, and the relative motion between the human and the robot) further complicate action recognition.

\vspace{-2mm}
Recognizing human actions from a robotic view presents several challenges, particularly when dealing with long distances and the interference of robotic motion. Traditional methods often fail to capture local details across varying scales and cannot effectively differentiate subtle human actions from kinematic disturbances caused by robot platform movements. To address these issues, we propose ACTIVE-PC, a comprehensive framework that utilizes Multilevel Neighborhood Sampling (MNS), Layered Recognizers (LR), and Elastic Ellipse Query (EEQ). MNS ensures both global coverage and retention of critical local details, while LR decouple features at different layers to preserve both fine-grained human actions and global kinematic variations. EEQ introduces axis-specific adaptive scaling to handle varying interaction distances, enhancing action differentiation by adjusting to the spatial distribution of motion. Experimental results demonstrate that this method improves action recognition performance in N-HRI scenarios.

\vspace{-4mm}
\section{Related Work}
\vspace{-1mm}
\subsection{Action Recognition Datasets}
Early 3D action recognition datasets, such as MSR-Action3D \cite{cvpr10msraction}, Multiview 3D Event \cite{iccv13wei2013modeling}, and NTU RGB+D \cite{NTU}, have played a crucial role in advancing action recognition technology. The MSR-Action3D dataset \cite{cvpr10msraction} primarily consists of sequences of gaming actions and was one of the first to incorporate depth information for action analysis. Many methods have been evaluated on this dataset, but due to the limited dataset size, recent approaches have achieved near saturation in accuracy. To address the limited sample size, the Multiview 3D Event dataset \cite{iccv13wei2013modeling} employs multiple depth sensors to capture multi-view representations of the same action. The NTU RGB+D dataset \cite{NTU}, the first large-scale RGB+D action recognition dataset, features a broader range of samples, more intra-class variations, and more camera views, capturing RGB video, depth sequences, and skeleton data. In recent years, research has shifted towards fine-grained action recognition tasks. The iMiGUE dataset \cite{imigue} aims to interpret athletes' emotional states by analyzing their micro-gestures, recording spontaneous micro-actions after a match. Similarly, the SMG dataset \cite{ijcv23chen2023smg} focuses on both micro-gesture and emotion recognition, documenting participants' micro-actions while narrating stories. However, action recognition for N-HRI involves more complex and diverse interaction states. Existing datasets typically rely on fixed relative positions and backgrounds, which are not ideal for action recognition in real-world human-machine interactions. In contrast, ACTIVE is the first large-scale action recognition dataset focused on N-HRI, capturing a variety of actions with disturbing motions in complex interaction scenarios.

\subsection{Action Recognition Methods}
RGB-based methods, such as TSN \cite{tsn}, VideoMAE \cite{videomae}, UniFormer \cite{UniFormer}, and InternVideo \cite{InternVideo}, have achieved high accuracy in general action recognition. Similarly, various methods have been developed for general action recognition from point cloud videos \cite{leaf,kinet,PointCMP}. To enhance the receptive field, P4Transformer \cite{p4trans} employs transformers to avoid explicit point tracking, although it does not fully encode the spatiotemporal structure. PST-Transformer \cite{psttrans} improves upon this by encoding features based on the spatiotemporal displacement between reference points and all points in the video. However, existing point cloud-based methods are typically limited to simple, single-action recognition and perform poorly on actions with disturbing motions. Our approach focuses on recognizing actions with disturbing motions in N-HRI, while remaining compatible with general action recognition. 

Currently, there is limited research specifically addressing micro-action recognition from point cloud videos. In RGB-based methods, Yonetani et al. \cite{yonetani} extracted hand-crafted features from both first-person and third-person videos for micro-action recognition. However, this approach requires both first-person and third-person videos, along with manual synchronization. Mi et al. \cite{mi2019} proposed a segment-level temporal pyramid for micro-action videos. Nonetheless, current micro-action recognition methods typically rely on fixed relative positions and backgrounds. In contrast, our approach adapts to different complex interaction scenarios, making it more suitable for action recognition in real-world human-machine interactions.

\section{Dataset}
\label{sec:3}
\subsection{The Action from Robotic View Dataset}

\begin{figure*}[!t]
	\centering 
	\begin{tabular}{c}		
		\includegraphics[width=17cm]{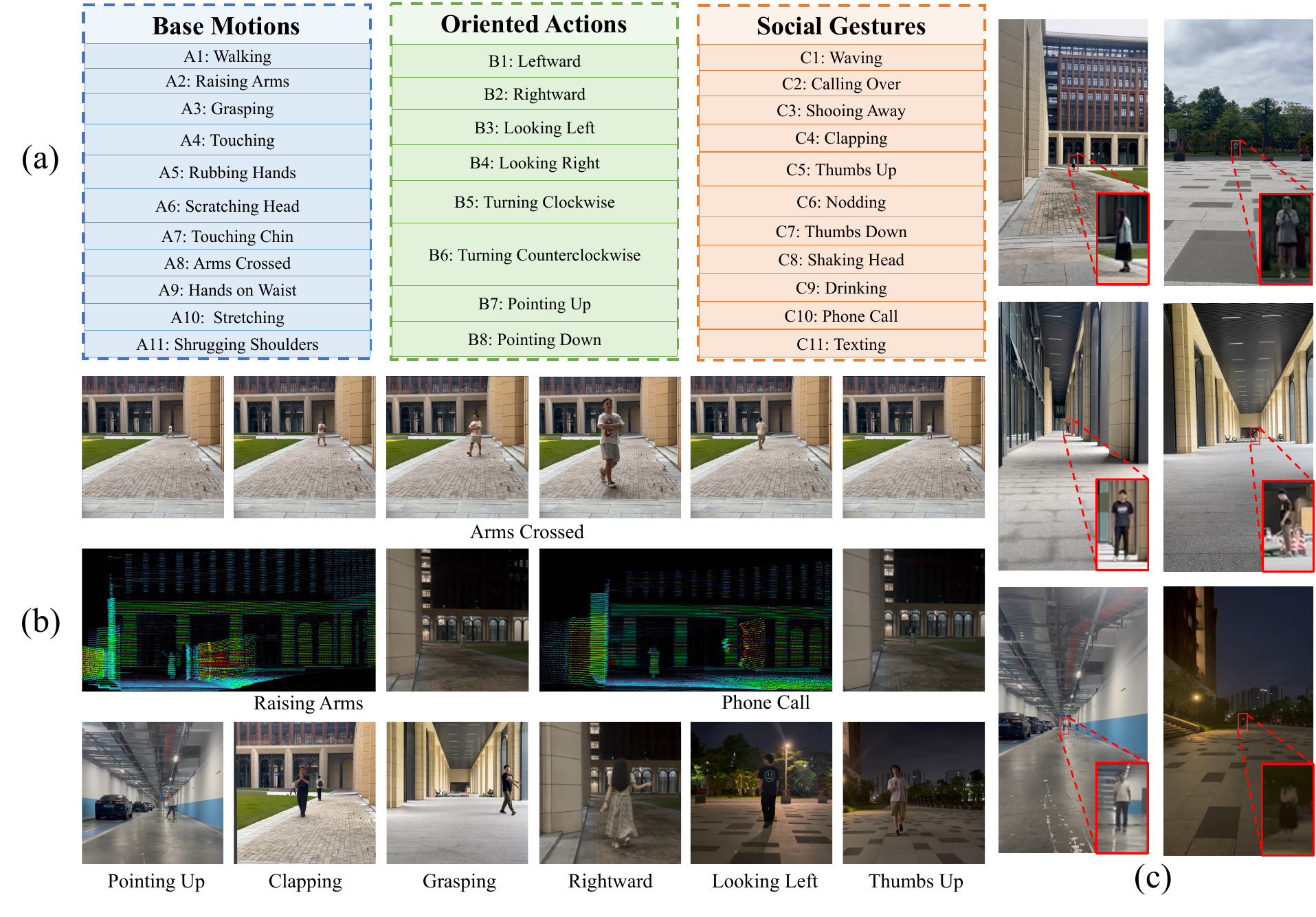}\\
	\end{tabular}%
     \vspace{-4mm}
	\caption{\textbf{Details of ACTIVE:} (a) Action Labels: Three categories are included: Base Motions, Oriented Actions, and Social Gestures. (b) The first row illustrates the significant variations in the captured results of the same action at different human-robot distances. The second row presents the raw point cloud captured by LiDAR and its corresponding RGB image. The point cloud modality retains more detailed geometric information, enabling long-range N-HRI understanding with strong robustness to interference. In contrast, the RGB modality complements this by providing rich texture and semantic information at close distances. The final row showcases additional action samples from our dataset. (c) Shooting Scenes: Six shooting scenes used in ACTIVE, covering both indoor and outdoor settings, as well as day and night conditions.}\label{fig:fig2}%
 \vspace{-6mm}
\end{figure*}%

ACTIVE is the first large-scale human behavior understanding dataset specifically designed for N-HRI scenarios (shown in \cref{fig:fig2}), featuring data from 80 participants along with annotated attribute information. It covers six distinct environments (including both indoor and outdoor settings, under varying daylight and nighttime conditions) and includes 30 action (\emph{e.g.}, Walking forward + Waving) categories. The dataset comprises 46,868 video instances, with an average duration of 3.2 seconds, totaling 41.7 hours. It includes two modalities: RGB and point cloud data. The point cloud data is captured using the HESAI AT128P LiDAR, with a horizontal field of view of 120°, a horizontal angular resolution of 0.1°, a vertical field of view of 25.4°, and a vertical resolution of 0.2°. The scanning frame rate is set at 10 Hz. The RGB video has a resolution of 1280 × 720 and a frame rate of 30 fps. \cref{tab:dataset1} compares ACTIVE dataset with existing human action recognition datasets. Below are some key features of ACTIVE.
\vspace{-4mm}
\paragraph{Diversity of Capture States and Behavior Patterns.} Traditional human behavior datasets typically focus on fixed capture modes and distances, often emphasizing a single behavior. In contrast, human behavior understanding in N-HRI requires handling more complex human-robot interaction scenarios. Therefore, ACTIVE collects data from subjects performing various actions under different human-robot states. In different settings, participants are instructed to perform multiple natural actions at varying distances. The linear distance between participants and the data capture platform ranges from 3 meters to 50 meters, with additional movement and vibrations to simulate changes in the robot’s state. Minor disturbances from vehicles and pedestrians are also allowed during recording.
\vspace{-4mm}
\paragraph{More Relevant Data Modalities.} Most existing human behavior datasets only provide traditional RGB video samples or depth modalities. While depth data can be converted into point clouds, it often lacks detailed geometric structure. Additionally, devices like Kinect suffer from short capture distances and poor performance under daylight conditions. In contrast, point cloud data captured by LiDAR retains richer geometric details, supports long-range N-HRI understanding, and is more robust to interference. The RGB modality complements this by offering rich texture and semantic information at close range. This underscores the importance of collecting diverse data modalities. Consequently, our varied data sources are designed to foster the development of robust models for human behavior analysis in complex N-HRI scenarios.
\vspace{-4mm}
\paragraph{Multiple Human Behavior Understanding Tasks.} To comprehensively analyze human behavior and actions in N-HRI, ACTIVE currently provides annotations for two primary tasks: human action recognition and human attribute recognition. Detailed annotations for the pose estimation task will be released at a later stage.

\subsection{Benchmark Evaluations}
\paragraph{Action Recognition.} To ensure consistency in benchmarking, we define the Cross-Subject Evaluation protocol and use classification accuracy (percentage) as the evaluation metric. A total of 80 participants were divided into a training group and a testing group, consisting of 53 and 27 subjects, respectively.
\vspace{-5mm}
\paragraph{Attribute Recognition.} For attribute recognition, we use the same dataset partitioning strategy as for action recognition. The performance of attribute recognition is evaluated by measuring the classification accuracy for each attribute.

\section{Methodologies}
The ACTIVE dataset captures natural human-robot interactions from robotic view, encompassing both large-scale global motion disturbances and subtle local postural changes in the human body. In this work, we primarily aim to address two key challenges posed by the ACTIVE dataset: (1) At long distances, human actions are subtle and rapid, making it a complex task to recognize these fine movements. (2) In dynamic interaction scenarios, large-scale global motion disturbances further increase the difficulty of action recognition. To address these challenges in ACTIVE, we introduce ACTIVE-PC. In \cref{sec:4.1.1}, we define the RARV task. In \cref{sec:4.1.2}, we present the pipeline of ACTIVE-PC. In \cref{sec:4.1.3} and \cref{sec:4.1.4}, we propose Multilevel Neighborhood Sampling (MNS) and Layered Recognizers (LR) to address challenge (1) and (2), and Elastic Ellipse Query to tackle challenge (2).

\subsection{Task Definition of RARV}\label{sec:4.1.1}
Consider an input video, \(\mathbf{X} = \{{I}_1, {I}_2, \dots, {I}_T\}\), where \(T\) is the length of the video. Each frame contains \(N\) points, and the video includes both human actions (\emph{e.g.}, raising arms) and kinematic disturbances (\emph{e.g.}, human-robot proximity and robot movement). Our goal is to filter out the influence of kinematic disturbances from the robot's perspective, in order to obtain the human action label \( Y \).

\subsection{Pipeline Overview}\label{sec:4.1.2}
\begin{figure*}[htbp]
	\centering 
	\begin{tabular}{c}		
		\includegraphics[width=17cm]{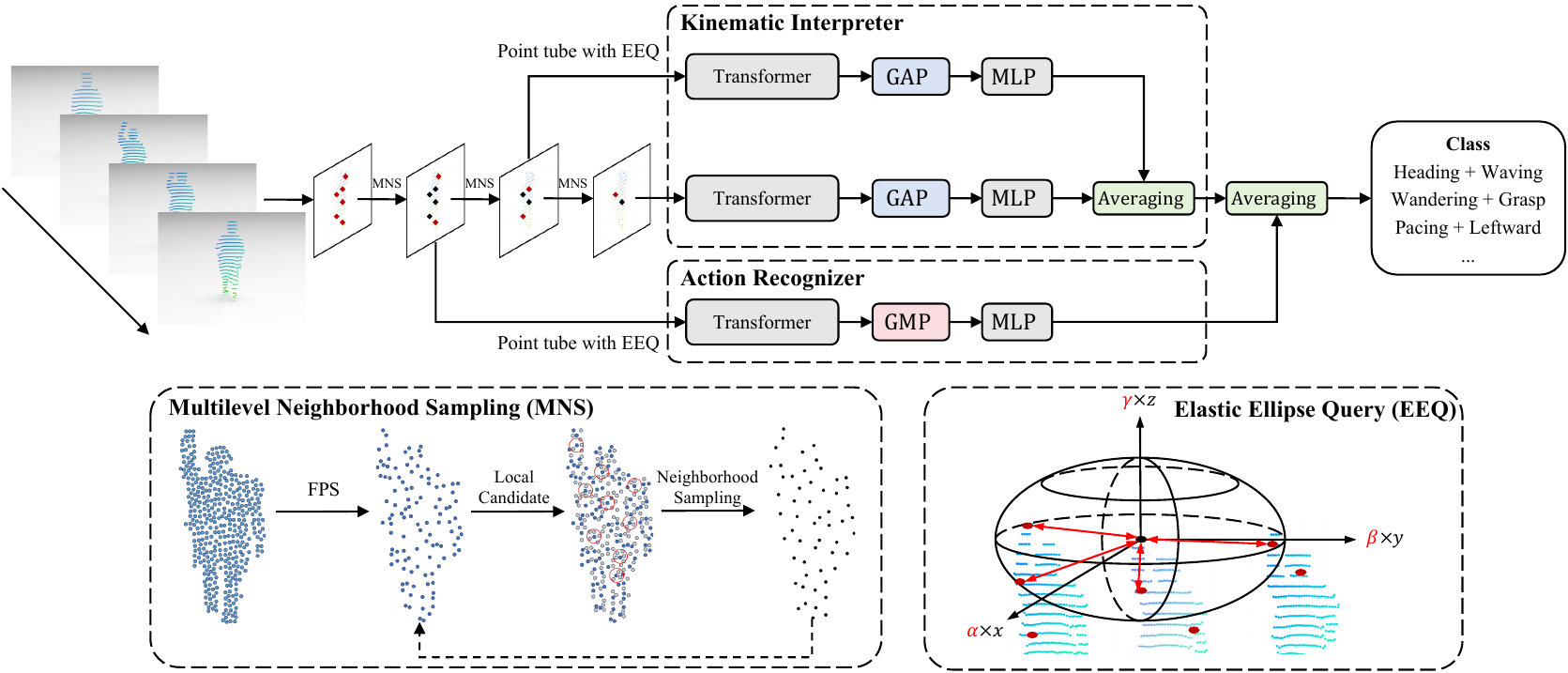}\\
	\end{tabular}%
	\caption{\textbf{ACTIVE-PC architecture.} First, the point cloud sequence undergoes Multilevel Neighborhood Sampling, followed by feature extraction using a point tube with Elastic Ellipse Query. The dense features are fed into Action Recognizer, while the sparse features are input into Kinematic Interpreter. The resulting human-action and kinematic scores are then fused to obtain the fine-grained action score.}\label{fig:pipeline1}%
    \vspace{-6mm}
\end{figure*}

The pipeline of ACTIVE-PC is shown in \cref{fig:pipeline1}. The input video \(\mathbf{X} \in \mathbb{R}^{T \times N \times 3}\) first undergoes MNS, producing point clouds with different densities: \(\mathbf{X_1} \in \mathbb{R}^{T/2 \times N/4 \times 3}\), \(\mathbf{X_2} \in \mathbb{R}^{T/2 \times N/16 \times 3}\) and \(\mathbf{X_3} \in \mathbb{R}^{T/2 \times N/32 \times 3}\). These point clouds, \(\mathbf{X_1}\), \(\mathbf{X_2}\) and \(\mathbf{X_3}\) are then processed through EEQ for adaptive neighbor querying, enabling spatiotemporal tracking and differentiation of human actions and kinematic disturbances, resulting in feature maps \(\mathbf{F}_1\), \(\mathbf{F}_2\) and \(\mathbf{F}_3\).

Low-density point clouds, which capture deeper features \(\mathbf{F}_2\) and \(\mathbf{F}_3\), are passed through the Kinematic Interpreter for human-robot state understanding. High-density point clouds, containing finer local details in \(\mathbf{F}_1\), are input to the Action Recognizer for human-action recognition. $\mathbf{Y}_{human}$ and $\mathbf{Y}_{kinematic}$ are fused to obtain the fine-grained action score \(\mathbf{Y}\).
The process is formulated as follows:
\begin{equation}
\resizebox{0.8\linewidth}{!}{$\displaystyle
\begin{aligned}
    \mathbf{X}_1,\mathbf{X}_2,\mathbf{X}_3 &= \mathbf{MNS}(\mathbf{X}),  \\
     \mathbf{F}_1,\mathbf{F}_2,\mathbf{F}_3 &=\mathbf{PointTube_{EEQ}}(\mathbf{X}_1,\mathbf{X}_2,\mathbf{X}_3),\\
      \mathbf{Y}_{kinematic} &= \mathbf{Interpreter_{kinematic}}(\mathbf{F}_2,\mathbf{F}_3),\\
        \mathbf{Y}_{human} &= \mathbf{Recognizer_{human}}(\mathbf{F}_1),\\
        \mathbf{Y} &= \frac{(\mathbf{Y}_{human} + \mathbf{Y}_{kinematic)})}{2}. \\
\end{aligned}$}
\end{equation}

\subsection{Multilevel Neighborhood Sampling}\label{sec:4.1.3}
To address the core challenges of the ACTIVE dataset, which include the precise perception of subtle human motions at long distances and the accurate decoupling of kinematic interference from human actions in dynamic environments, Multilevel Neighborhood Sampling (MNS, \cref{sec:4.1.3}) and Layered Recognizers (LR, \cref{sec:4.1.4}) work in tandem.

Traditional methods typically apply multi-scale or different sampling rates of farthest point sampling (FPS) directly to the entire point cloud. While this approach ensures globally uniform sampling, it often neglects the correlation between local details across different scales, which is critical in the ACTIVE dataset. In situations where the human-robot distance is large, high-density local sampling can better capture small yet distinguishable motion features.

To address this, we propose Multilevel Neighborhood Sampling (MNS), which builds hierarchical dependencies using local neighborhood constraints, ensuring both global coverage and retention of local details. The implementation of MNS is shown in \cref{fig:pipeline1}. The black points represent the anchor points obtained through downsampling, while the red points represent the neighboring points. The process is as follows:
First, for the input point cloud video data \(\mathbf{X} \in \mathbb{R}^{T \times N \times 3}\), we apply FPS to sample \(N_1\) points globally, obtaining the first layer of anchor points: 
\begin{equation}
\mathbf{X}_1 = \operatorname{FPS}(\mathbf{X}; N_1).\\
\end{equation}

For the \(l\)-th layer, instead of directly sampling globally, we construct a candidate set within the local neighborhood of the previous layer's sampled points \(\mathbf{x}_{l-1,j}\) (where \(j=1, \dots, N_{l-1}\)). Using a radius \(r\) and a candidate number \(k\), we define the spherical neighborhood as:
\begin{equation}
\mathcal{N}( \mathbf{x}_{l-1,j} ) = \mathbf{x}_i : |\mathbf{x}_i - \mathbf{x}_{l-1,j}|_2 \leq r^{(l)}, |\mathcal{N}| \leq k^{(l)}.\\
\end{equation}
Then, the union of the neighborhoods across all \(j\) is taken and duplicates are removed, forming the local candidate set: 
\begin{equation}
\widehat{\mathbf{X}}_l = \bigcup \mathcal{N}(\mathbf{x}_{l-1,j}).\\
\end{equation}

Through MNS, we can capture global structure in the sparse layers (counteracting the kinematic interference caused by platform motion from robotic view), while fine-tuning the extraction of subtle human motion changes in the denser layers. This approach precisely aligns with the ACTIVE dataset’s high demand for fine-grained local information when capturing complex human-robot dynamics.

\subsection{Layered Recognizers}
\label{sec:4.1.4}
Considering the characteristics of the ACTIVE dataset, human action recognition requires not only capturing fine-grained, small-scale movements but also filtering out global kinematic variations caused by robot platform motion. To address these challenges, we design Layered Recognizers, which decouples features from different layers by sending them through specialized sub-networks. The core of the design includes two branches:
\paragraph{Kinematic Interpreter:} For features obtained from larger sampling scales (\emph{i.e.}, sparser layers, such as \(\mathbf{F}_2\) and \(\mathbf{F}_3\)), this branch focuses on global kinematic impacts caused by platform motion and changes in the human-robot spatial relationship from the robot's perspective. Since these global features are relatively dispersed in the temporal dimension, we use Global Average Pooling (GAP) to effectively aggregate information from all spatiotemporal regions, ensuring that contributions from different regions are considered. The process is as follows:
\begin{equation}
\resizebox{0.85\linewidth}{!}{$\displaystyle 
\begin{aligned}
\mathbf{Y}_{kinematic} &= \mathbf{Interpreter_{kinematic}}(\mathbf{F}_2,\mathbf{F}_3)\\
&= \frac{1}{2} \left( \mathbf{MLP}(\mathbf{GAP}(\mathbf{Transformer}(\mathbf{F}_2))) \right. \\
& \quad + \left. \mathbf{MLP}(\mathbf{GAP}(\mathbf{Transformer}(\mathbf{F}_3))) \right).\\
\end{aligned}$}
\end{equation}

\paragraph{Human Action Recognizer:} For features extracted from lower layers with denser point clouds (\emph{e.g.}, \(\mathbf{F}_1\)), these features more directly reflect subtle human actions. Here, we use Global Max Pooling (GMP) to select the most significant spatiotemporal points, emphasizing key fine-grained action details. The calculation is as follows:
\begin{equation}
\begin{aligned}
\mathbf{Y}_{human}  &= \mathbf{Recognizer_{human}}(\mathbf{F}_1)\\
&= \mathbf{MLP}(\mathbf{GMP}(\mathbf{Transformer}(\mathbf{F}_1)).\\
\end{aligned}
\end{equation}
Finally, we fuse the classification scores from both branches, averaging their contributions. This approach retains the robustness of global kinematic interpretation while capturing the high discriminability of subtle action changes.

\subsection{Elastic Ellipse Query}\label{sec:4.1.5}
Let the 4D point cloud set be \( P = \{p_1, p_2, \dots, p_N\} \), where each point \( p_n = (x_n, y_n, z_n) \) represents the neighboring point \( n \) of a query point. Given a query point \( p_q \) and a radius \( r \), the objective of the Ball Query is to find all points whose distance from the query point \( p_q \) is less than or equal to \( r \).
Let \( d(p_n, p_q) \) denote the Euclidean distance between point \( p_n \) and query point \( p_q \):
\begin{equation}
\scalebox{0.9}{$d(p_n, p_q) = \sqrt{(x_n - x_q)^2 + (y_n - y_q)^2 + (z_n - z_q)^2}$},
\end{equation}
where \( p_q = (x_q, y_q, z_q) \) is the query point.

The goal of Ball Query is to identify all points \( p_n \) satisfying the following condition:
\begin{equation}
\mathcal{B}(p_q, r) = \{ p_n \mid d(p_n, p_q) \leq r \}.
\end{equation}
That is, the set of points within the spherical neighborhood \( \mathcal{B}(p_q, r) \) consists of all points \( p_n \) such that \( d(p_n, p_q) \leq r \).

In N-HRI, conventional spherical neighborhood queries (\eg, Ball Query) exhibit inherent limitations when confronted with kinematic-induced geometric distortions. The relative motion between mobile robots and humans introduces anisotropic deformations in the spatiotemporal point cloud distribution. Three critical challenges emerge: 1) Scale Discrepancy: Human actions (\eg, hand gestures) manifest as localized deformations (0.1-0.5m) within the robot's egocentric coordinates, while kinematic disturbances induce macroscopic displacements (up to 50m) along the xy-plane. 2) Axis-wise Heterogeneity: Vertical displacements (along the z-axis) primarily reflect genuine human joint movements, while horizontal movements (in the xy-plane) are more susceptible to artifacts caused by the robot's self-motion and changes in the human-robot spatial positioning. 3) Dynamic Occlusion: Transient point cloud density variations caused by viewpoint changes violate the isotropic neighborhood assumption.

For RARV, visual similarities influenced by kinematic disturbances from robotic view complicate the differentiation, causing deformations in human-actions and irregular distributions in the spatiotemporal domain. Unlike kinematic disturbances, which have a large range of movement, human-actions are constrained to smaller areas. For instance, while walking involves significant movement in the xy-plane, human-actions like waving hands are restricted to a smaller range. Due to the influence of kinematic, points with large xy distance at different times are more likely to represent the same point, while points with large differences in the vertical dimension are more likely to be distinct.

To address these challenges, we propose the \textit{Elastic Ellipse Query}. Considering the distribution differences between datasets, we introduce axis-specific adaptive query scale learning. This mechanism establishes an anisotropic metric space through learnable axis-specific scaling, enabling partial tracking of points and differentiation of actions. Given a query point \( p_q \) and a neighborhood point \( p_n \), our adaptive distance metric is formulated as follows:
\begin{equation}
\scalebox{0.8}{$d(p_n, p_q) = \sqrt{\alpha(x_n - x_q)^2 + \beta(y_n - y_q)^2 + \gamma(z_n - z_q)^2}$},
\end{equation}
where \( \Omega = \{\alpha, \beta, \gamma\} \) are trainable scaling parameters that automatically adjust the neighborhood aspect ratios. For the ACTIVE dataset, the final values of \( \alpha \), \( \beta \), and \( \gamma \) are $(3.5632, 3.6789, 2.8038)$, consistent with the expected scale ratios. This elliptic metric space provides following distinctive advantages: 1) Kinematic-Aware Adaptation: The horizontal scaling factors (\(\alpha, \beta\)) suppress planar displacements induced by the robot while preserving fine-grained human motions. In our experiments with the ACTIVE dataset, the optimized parameters (\(\alpha=3.56\), \(\beta=3.68\)) effectively compress the neighborhood range in the xy-plane relative to the z-axis (\(\gamma=2.80\)). The vertical scaling factor \(\gamma\) remains sensitive to subtle posture changes, such as a 0.2-meter arm elevation, while the horizontal scaling factors adaptively filter out irrelevant positional shifts. 2) Dynamic Focus Adjustment: Unlike fixed elliptical kernels, our parameters evolve during training to handle varying interaction distances (3-50m).

\section{Experiment}
\begin{table}[t!]
\centering
\caption{Action recognition accuracy (\%) of ACTIVE-PC on ACTIVE under different dataset settings.}
\renewcommand{\arraystretch}{1}
  \footnotesize
\begin{tabular*}{\columnwidth}{c|c|c@{\extracolsep{\fill}}c@{\extracolsep{\fill}}c@{\extracolsep{\fill}}}
        \toprule
       \multirow{2}{*}{Method}                 & \multirow{2}{*}{Venue}            &\multicolumn{3}{c}{Setting}  \\
       \cline{3-5}
       ~&~& 8 $\times$ 512  &12 $\times$ 512&12 $\times$ 768  \\
        \cline{1-5}
        PSTNet \cite{pstnet}&ICLR'21          & 18.85      & 26.49&34.10\\                   
        \cline{1-5}
        P4Transformer \cite{p4trans}&CVPR'21    & 37.97  & 40.56&43.88  \\                                   
        \cline{1-5}   
      PSTNet++ \cite{pstnet++} & TPAMI’22           &34.98    &38.81&  40.29  \\
        \cline{1-5}      
        PST-Transformer \cite{psttrans} & TPAMI’23  & 41.61 & 42.92&46.76\\
        \cline{1-5}    
        3DinAction \cite{3DInAction}&CVPR'24    & 44.88 &  46.23&49.19\\    
        \cline{1-5}   
        ACTIVE-PC w/o EEQ &-  &52.29& 53.32& 55.11\\
        \cline{1-5}   
        ACTIVE-PC (Ours) &-  & \textbf{58.71}& \textbf{59.18} & \textbf{60.10}\\
        \bottomrule
    \end{tabular*}
\renewcommand{\arraystretch}{1}
\label{tab:2}
\vspace{-2mm}
\end{table}

\begin{table}[t!]
\centering
\caption{Action recognition accuracy (\%) of ACTIVE-PC on NTU RGB+D. CS and CV represent the cross-subject and cross-view evaluation protocols, respectively.}
\renewcommand{\arraystretch}{0.7}
\vspace{-2mm}
\footnotesize
\begin{tabular*}{\columnwidth}{c@{\extracolsep{\fill}}c@{\extracolsep{\fill}}c@{\extracolsep{\fill}}c@{\extracolsep{\fill}}}
\toprule
Method& Venue & CS & CV  \\
\midrule[0.3mm]
PSTNet  \citep{pstnet} & ICLR'2021& 90.5 & 96.5  \\
P4Transformer \citep{p4trans}&CVPR'2021& 90.2 & 96.4  \\
PointCMP \citep{PointCMP} &CVPR'2023& 88.5 & - \\
PointCPSC \citep{PointCP}& ICCV'2023& 88.0 & -\\
PST-Transformer  \citep{psttrans}&TPAMI'2023 & 91.0 & 96.4 \\
\hline
\addlinespace[0.5em]
ACTIVE-PC (Ours)&- &\textbf{91.7} & \textbf{96.8} \\ 
\bottomrule
\end{tabular*}
\label{tab:ntu}
\vspace{-2mm}
\end{table}

\begin{table}[t!]
\centering
\caption{Action recognition accuracy (\%) of ACTIVE-RGB on ACTIVE under different dataset settings.}
\renewcommand{\arraystretch}{1}
 \footnotesize
\begin{tabular*}{\columnwidth}{c|c|@{\extracolsep{\fill}}cc}
        \toprule
        \multirow{2}{*}{Method}                   & \multirow{2}{*}{Venue}            & \multicolumn{2}{c}{Setting}  \\
        \cline{3-4}
        ~&~& 8 $\times$ 224  &12 $\times$ 224 \\
        \cline{1-4}
        VideoSwin \cite{videoswin}&CVPR'22         & 41.07                & 44.20 \\    
       \cline{1-4}
        MViT V2 \cite{mvitv2}&CVPR'22          & 38.15               & 41.83\\           
       \cline{1-4}   
           VideoMAE \cite{videomae} & NeurIPS'22                    & 42.46               & 45.63 \\
        \cline{1-4} 
        UniFormer \cite{UniFormer}&TPAMI'23                & 42.55                &  45.60\\                                                                
       \cline{1-4}
        VideoMAE V2 \cite{videomaev2} &CVPR'23                    & 45.39               &47.98 \\
       \cline{1-4}  
        UniFormerv2 \cite{UniFormerv2} &ICCV'23      & 48.21                & 50.74 \\
        \cline{1-4}
        InternVideo2 \cite{internvideo2} &ECCV'24      & 52.52      & 54.88 \\
        \cline{1-4}
        ACTIVE-RGB (Ours) &-       & \textbf{55.90}             & \textbf{57.56}  \\
        \bottomrule
    \end{tabular*}
\renewcommand{\arraystretch}{1}
\label{tab:4}
\vspace{-2mm}
\end{table}

In this section, we evaluate the performance of state-of-the-art methods and ACTIVE-PC, on the ACTIVE dataset across both point cloud and RGB modalities. We also conduct ablation experiments on the proposed components.

\subsection{Metrics and Implementation Details}
For evaluation, we use standard action recognition metrics, specifically classification accuracy. In the point cloud modality, we first apply the TransFusion\cite{transfusion} method to detect human subjects within the scene. The point cloud sequence is then segmented into fixed-length clips, with each frame sampled to 768 points. For training, video-level labels are assigned as clip-level labels. The color information of the points is not used. For evaluation, the video-level prediction is obtained by averaging the clip-level predicted probabilities. The model is trained using the SGD optimizer for 50 epochs, with a learning rate of 0.01, decaying by a factor of 0.1 at epochs 20 and 30. For the RGB modality, human bodies are detected using the object detection method \cite{yolov10}. Each video is evenly split into 8 or 12 segments, with one frame randomly sampled from each segment. Each frame is resized to 224 × 224. We use the AdamW optimizer \cite{Adamw} for 50 epochs, with a learning rate of 0.0002. All experiments are conducted on the PyTorch platform, utilizing an NVIDIA RTX 4090 GPU.

\begin{table}[t!]
    \centering
    \caption{Impact of Layered Recognizers, MNS and EEQ of ACTIVE-PC on ACTIVE.}
    \renewcommand{\arraystretch}{0.6}
    \vspace{-2mm}
 \footnotesize
\begin{tabular*}{\columnwidth}{c@{\extracolsep{\fill}}c@{\extracolsep{\fill}}c@{\extracolsep{\fill}}c@{\extracolsep{\fill}}}
    \toprule
    Layered Recognizers & MNS  & EEQ & Accuracy (\%) \\ 
    \midrule
    \ding{55}&  \ding{55}& \ding{55}& 46.76 \\ 
    \ding{51} & \ding{55}& \ding{55}&   51.98   \\ 
    \ding{51} & \ding{51}  &\ding{55}&   55.11  \\ 
    \ding{51}&  \ding{55} &\ding{51}&   55.63  \\ 
    \ding{51} & \ding{51}  &\ding{51}&   60.10  \\ 
    \bottomrule
    \end{tabular*}
    \vspace{-2mm}
    \label{tab:3}
\end{table}

\begin{table}[t!]
\centering
\caption{Influence of the spatial radius $r$. EEQ reduces the sensitivity to the hyperparameter $r$.}
\renewcommand{\arraystretch}{1.2}
\vspace{-3mm}
  \footnotesize
\begin{tabular*}{\columnwidth}{cc@{\extracolsep{\fill}}c@{\extracolsep{\fill}}c@{\extracolsep{\fill}}c@{\extracolsep{\fill}}c@{\extracolsep{\fill}}c@{\extracolsep{\fill}}c}
        \toprule
       \multirow{2}{*}{EEQ}     &\multicolumn{6}{c}{Spatial radius $r_s$} &\multirow{2}{*}{Variance} \\
       \cline{2-7}
       ~& 0.1  &0.2&0.5&1&1.5&2 \\
        \cline{1-8}
         \ding{55}& 54.83&55.11&  55.39& 54.52&54.21&53.51&0.456\\
        \cline{1-8}   
        \ding{51} &59.98&60.10&60.04&59.73&59.54&59.30&0.100\\
        \bottomrule
    \end{tabular*}
\renewcommand{\arraystretch}{1}
\label{tab:eeq}
\vspace{-2mm}
\end{table}

\begin{figure}[t!]
	\centering 
	\begin{tabular}{c}		
		\includegraphics[width=8cm]{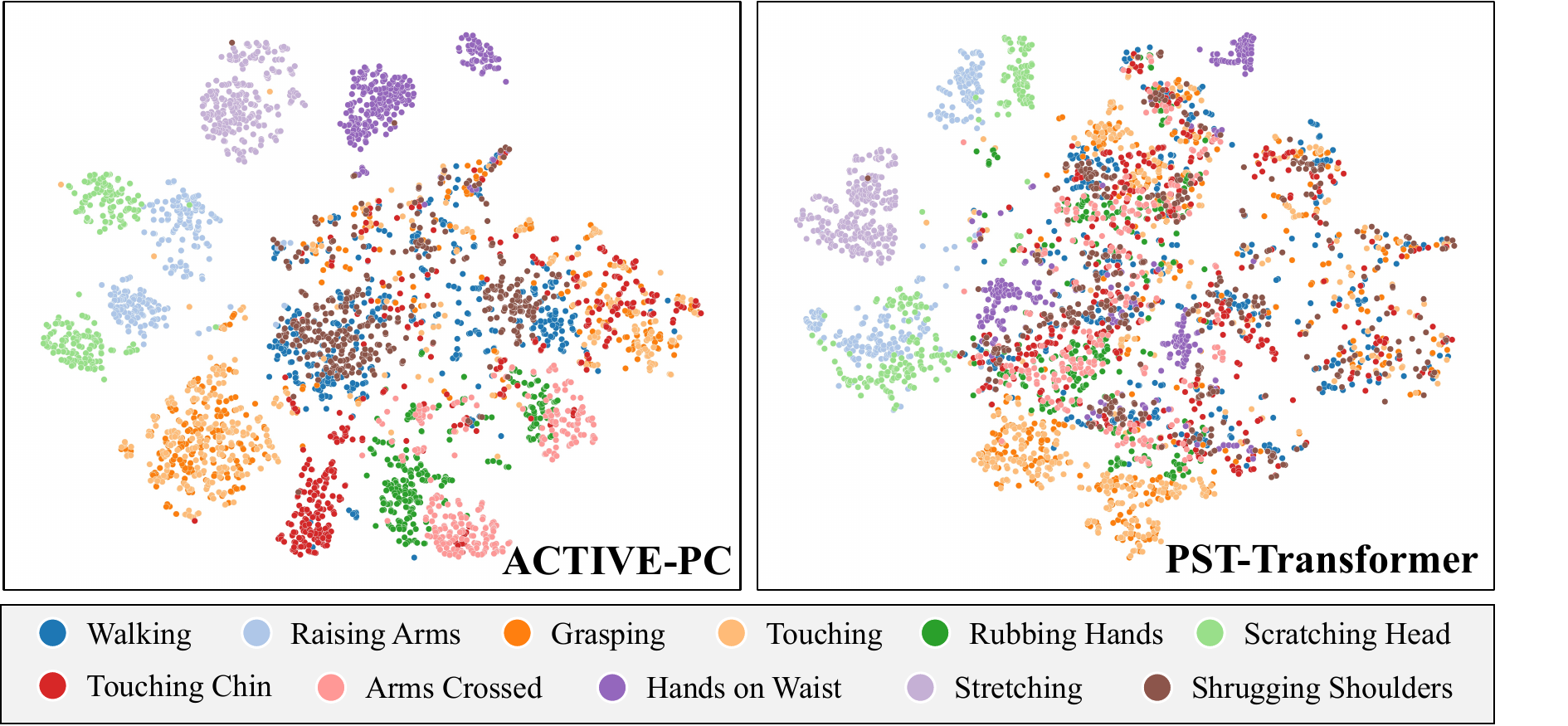}\\
	\end{tabular}%
    \vspace{-2mm}
	\caption{t-SNE distribution of PST-Transformer \cite{p4trans} and ACTIVE-PC on ACTIVE dataset. Different colors represent different action categories, with each point corresponding to a video sample. All 11 categories from the Base Motions are visualized.}\label{fig:fig5}%
    \vspace{-6mm}
\end{figure}%

\subsection{Point Cloud Modality Evaluation}
We first validate the performance of ACTIVE-PC on the ACTIVE dataset. As shown in \cref{tab:2}, Transformer-based models, such as P4Transformer and PST-Transformer, outperform CNN-based models achieving Acc-Top1 scores of 43.88\% and 46.76\%, respectively. ACTIVE-PC surpasses PST-Transformer by 13.34\%, reaching an Acc-Top1 of 60.10\%, demonstrating the effectiveness of our method. To validate the robustness of the model, we conduct experiments on two general action recognition datasets, MSR-Action3D and NTU RGB+D. \cref{tab:ntu} presents the results on NTU RGB+D.

\subsection{RGB Modality Evaluation}
ACTIVE-RGB is built upon the ViT \cite{Dosovitskiy2020AnII}, augmented with Layered Recognizers. \cref{tab:4} presents a performance comparison of ACTIVE-RGB with other state-of-the-art methods in the RGB modality. ACTIVE-RGB outperforms the leading method, the baseline model InternVideo2, by 2.68\%, validating the effectiveness of our approach.

\subsection{Ablation Study}
\paragraph{Evaluation of different Components:}  We conduct ablation experiments on the key components of our model, as well as the number of sampled frames per point cloud video. Results in \cref{tab:3} show that each proposed modification contributes positively. Specifically, w/o EEQ and w/o MNS show performance drops of 4.99\% and 4.47\%, respectively, compared to ACTIVE-PC. The significant performance degradation observed with w/o Layered Recognizers highlights the effectiveness of fusing high-frequency and low-frequency features. Allowing the model to simultaneously focus on both action and kinematic variations enhances its ability to handle complex environments. 
\vspace{-2mm}
\paragraph{Evaluation of the Stability of EEQ:}
\cref{tab:eeq} compares the performance of the full ACTIVE-PC with its variant, which excludes the EEQ module, under different neighborhood radius settings. Traditional neighborhood query methods are constrained by fixed neighborhood radii, whereas EEQ adapts by learning the intrinsic geometric properties of the data distribution, allowing for flexible scaling of the spherical neighborhood, thereby enhancing robustness. The variance of accuracy (acc) for ACTIVE-PC across different neighborhood radii is 0.100, significantly outperforming the variant without EEQ, which exhibits a variance of 0.456.

\subsection{Visualization and Analysis}
We utilize t-SNE \cite{tsne} visualization to analyze feature distributions from the point cloud modality, as shown in \cref{fig:fig5}. ACTIVE-PC distinctly separates similar actions (e.g., "Raising Arms" vs. "Scratching Head" and "Arms Crossed" vs. "Hands on Waist"), whereas PST-Transformer \cite{p4trans} produces more overlapping clusters. This demonstrates ACTIVE-PC's superior ability to differentiate subtle action categories from a robotic viewpoint.

\section{Conclusion}
To the best of our knowledge, our dataset is the first human action recognition dataset specifically designed for Natural Human-Robot Interaction scenarios, laying the foundation for more accurate and detailed understanding of human actions in N-HRI. We also propose ACTIVE-PC designed for action recognition in N-HRI scenarios. Experimental results demonstrate the effectiveness of this approach.

\noindent\textbf{Acknowledgements} This work was supported by National Natural Science Foundation of China (No. 62473007), Natural Science Foundation of Guangdong Province (No. 2024A1515012089), Shenzhen Innovation in Science and Technology Foundation for The Excellent Youth Scholars (No. RCYX20231211090248064).

{
    \small
    \bibliographystyle{ieeenat_fullname}
    \bibliography{main}
}

\end{document}